\documentclass[runningheads,a4paper]{llncs}
\usepackage{amssymb}
\usepackage{graphicx}

\usepackage{amsmath}
\usepackage{hyperref}
\usepackage{booktabs}
\usepackage{multirow}

\usepackage{mathptmx}
\usepackage[10pt]{moresize}

\usepackage{url}
\urldef{\mailsa}\path|{fady.medhat, david.chesmore, john.robinson}@york.ac.uk|    
\newcommand{\keywords}[1]{\par\addvspace\baselineskip
\noindent\keywordname\enspace\ignorespaces#1}

\begin{document}

\mainmatter  

\title{Masked Conditional Neural Networks\\ for 
Audio Classification}

\titlerunning{Masked Conditional Neural Networks for Audio Classification}

%
%
\author{Fady Medhat
\and David Chesmore\and John Robinson}
\authorrunning{F. Medhat, D. Chesmore, J. Robinson}
\institute{Department of Electronic Engineering\\ University of York, York\\ United Kingdom\\
\mailsa}

%
%
\toctitle{Lecture Notes in Computer Science}
\tocauthor{Authors' Instructions}
\maketitle
\begin{abstract}
We present the ConditionaL Neural Network (CLNN) and the Masked ConditionaL Neural Network (MCLNN)\footnote[1]{Code: https://github.com/fadymedhat/MCLNN}  designed for temporal signal recognition. The CLNN takes into consideration the temporal nature of the sound signal and the MCLNN extends upon the CLNN through a binary mask to preserve the spatial locality of the features and allows an automated exploration of the features combination analogous to hand-crafting the most relevant features for the recognition task. MCLNN has achieved competitive recognition accuracies on the GTZAN and the ISMIR2004 music datasets that surpass several state-of-the-art neural network based architectures and hand-crafted methods applied on both datasets.     
\end{abstract}

\section{Introduction}
The success of the deep neural network architectures in image recognition \cite{RN367} induced applying these models for audio recognition \cite{RN44}\cite{RN431}. One of the main drivers for the adaptation is the need to eliminate the effort invested in hand-crafting the features required for classification. Several neural networks based architectures have been proposed, but they are usually adapted to sound from other domains such as image recognition. This may not exploit sound related properties. The Restricted Boltzmann Machine (RBM)\cite{RN290} treats sound as static frames ignoring the inter-frame relation and the weight sharing in the vanilla Convolution Neural Networks (CNN)\cite{RN307} does not preserve the spatial locality of the learned features, where limited weight sharing was proposed in \cite{RN44} in an attempt to tackle this problem for sound recognition.
\begin{figure}
\begin{center}
\centerline{\includegraphics[width=6.0cm]{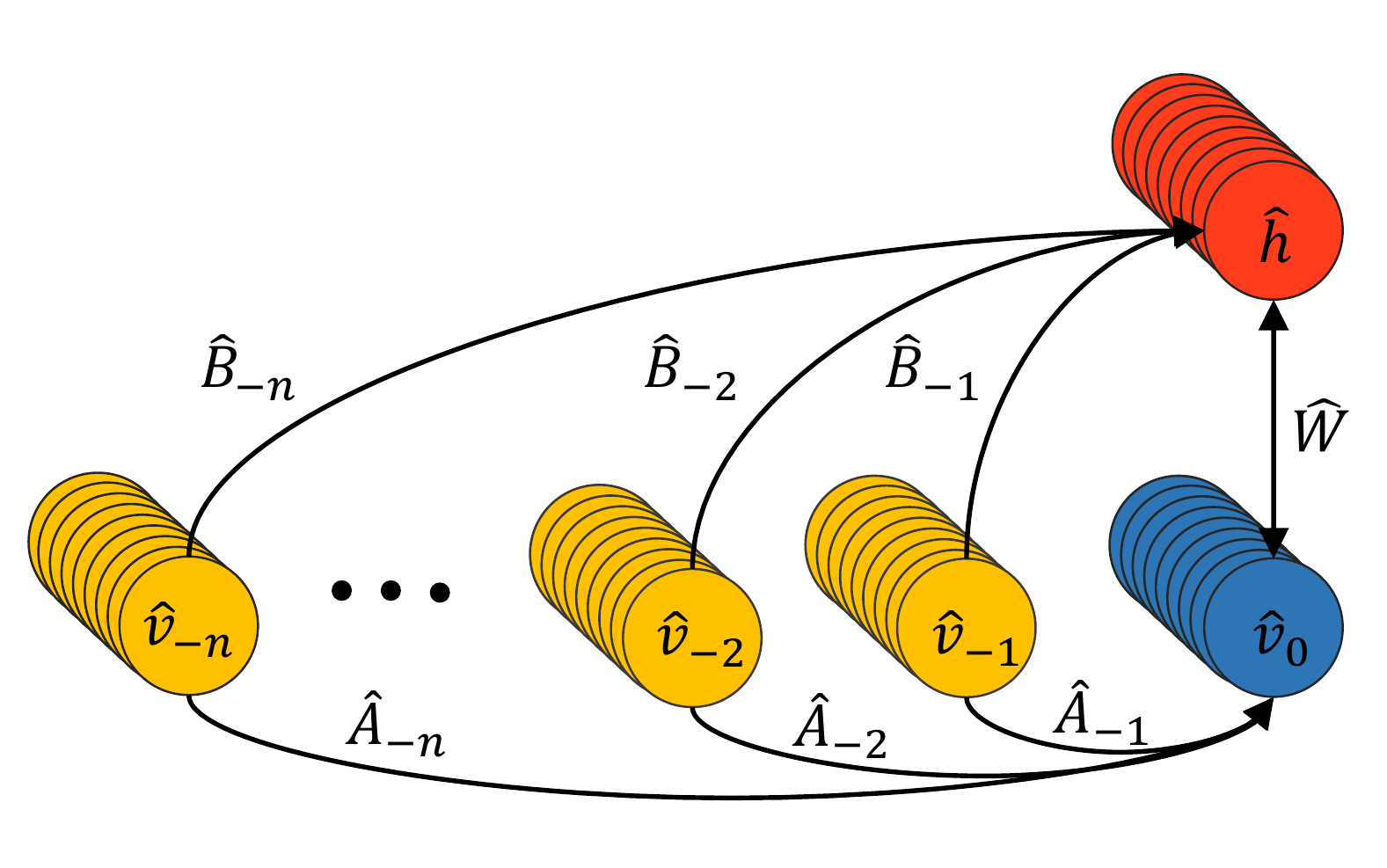}}
\vskip -0.1in
\caption{Conditional Restricted Boltzmann Machine}
\label{fig:crbm}
\end{center}
\vskip -0.4in
\end{figure}\\

The Conditional Restricted Boltzmann Machine (CRBM) \cite{RN207} in Fig. \ref{fig:crbm} extends the RBM \cite{RN414} to the temporal dimension. This is applied by including conditional links from the previous frames $(\hat{v}_{-1},\hat{v}_{-2},..., \hat{v}_{-n})$ to both the hidden nodes $\hat{h}$ and the current visible nodes $\hat{v}_{0}$ using the links $(\hat{B}_{-1},\hat{B}_{-2},..., \hat{B}_{-n})$  and the autoregressive links $(\hat{A}_{-1}, \hat{A}_{-2},..., \hat{A}_{-n})$, respectively as depicted in Fig. \ref{fig:crbm}. The Interpolating CRBM (ICRBM) \cite{RN257} achieved a higher accuracy compared to the CRBM for speech phoneme recognition by extending the CRBM to consider both the previous and future frames.

The CRBM behavior (and similarly this work) overlaps with that of a Recurrent Neural Network (RNN) such as the Long Short-Term Memory (LSTM) \cite{RN348}, an architecture designed for sequence labelling. The output of an RNN at a certain temporal instance depends on the current input and the the hidden state of the network's internal memory from the previous input. Compared to an LSTM, a CRBM does not require an internal state, since the influence of the previous temporal input states is concurrently considered with the current input. Additionally, increasing the order $n$ does not have the consequence of the vanishing or exploding gradient related to the Back-Propagation Through Time (BPTT) as in recurrent neural networks that LSTM was introduced to solve, since the back-propagation in a CRBM depends on the number of layers as in normal feed-forward neural networks. 

Inspired by the human visual system, the Convolutional Neural Network (CNN) depends on two main operations namely the convolution and pooling. In the convolutional operation, the input (usually a 2-dimensional representation) is scanned (convolved) by a small-sized weight matrix, referred to as a filter. Several small sized filters, e.g. $5\times5$, scan the input to generate a number of feature maps equal to the number of filters scanning the input. A pooling operation generates lower resolution feature maps, through either a mean or a max pooling operation. CNN depends on weight sharing that allows applying it to images of large sizes without having a dedicated weight for each pixel, since similar patterns may appear at different locations within an image. This is not optimally suitable for time-frequency representations, which prompted attempts to tailor the CNN filters for sound \cite{RN44},\cite{RN340},\cite{RN377}.

\section{Conditional Neural Networks}
\vskip -0.1in
\begin{figure}
\vskip 0in
\begin{center}
\centerline{\includegraphics[width=8cm]{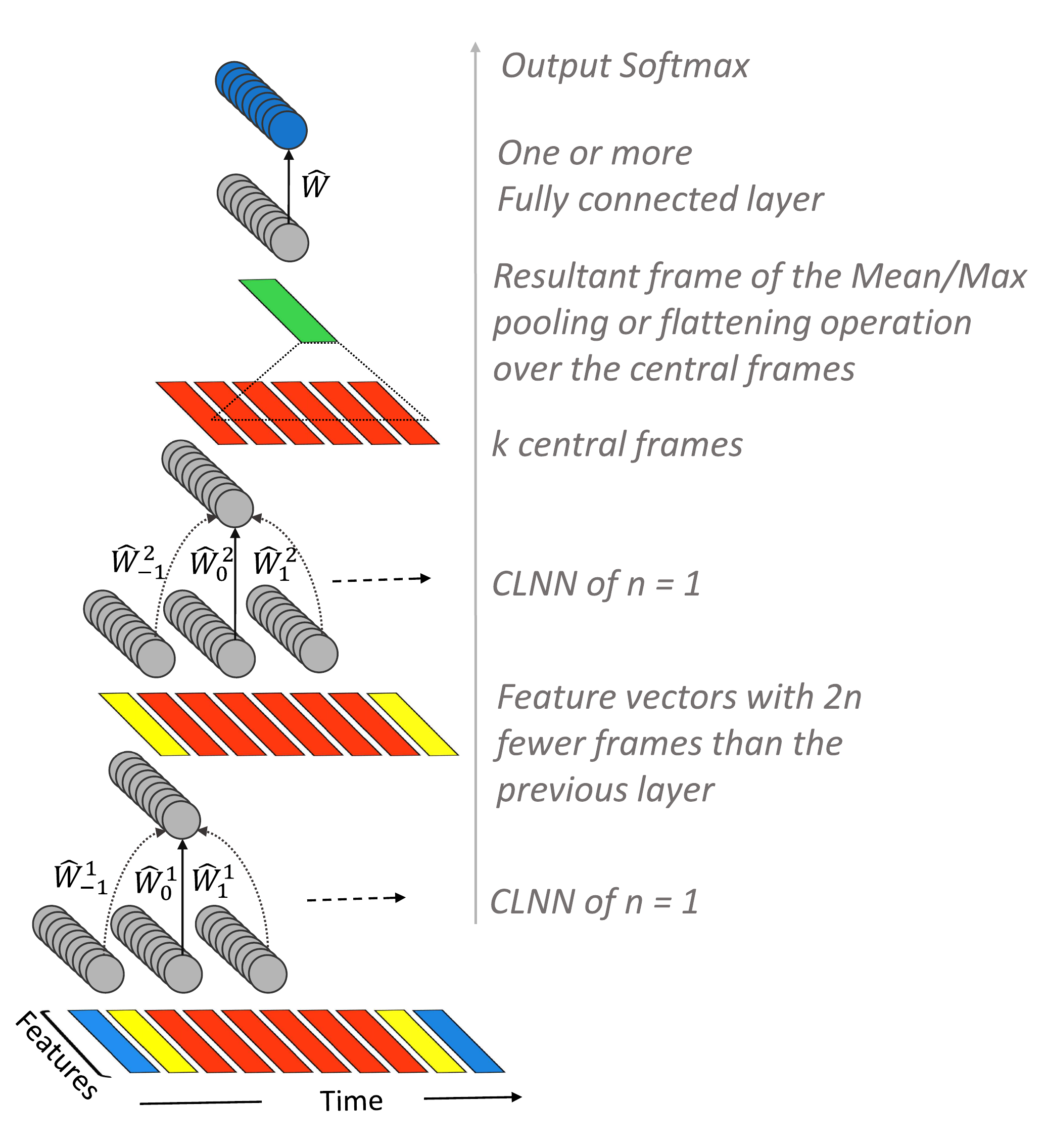}}
\vskip -0.1in
\caption{Two CLNN layers with $n=1$.}
\label{fig:clnn}
\end{center}
\vskip -0.5in
\end{figure} 

In this work, we introduce the ConditionaL Neural Network (CLNN). The CLNN adaptes from the Conditional RBM the directed links between the previous visible and the hidden nodes and extends to future frames as in the ICRBM. Additionally, the CLNN adapts a global pooling operation \cite{RN332}, which behaves as an aggregation operation found to enhance the classification accuracy in \cite{RN295}.
The CLNN allows the sequential relation across the temporal frames of a multi-dimensional signal to be considered collectively by processing a window of frames. The CLNN has a hidden layer in the form of a vector having e neurons, and it accepts an input of size $[d, l]$, where $l$ is the feature vector length and $d=2n+1$ ($d$ is the number of frames in a window, $n$ is the order for the number of frames in each temporal direction and 1 is for the window's middle frame). Fig. \ref{fig:clnn} shows two CLNN layers each having an order $n = 1$, where $n$ is a tunable hyper-parameter to control the window's width. Accordingly, each CLNN layer in the figure has a 3-dimensional weight tensor composed of one central matrix $\hat{W}_0^m$ and two off-center weight matrices, $\hat{W}_{-1}^m$ and $\hat{W}_{1}^m$ ($m$ is the layer id). During the scanning of the signal across the temporal dimension, a frame in the window at index \textit{u} is processed with its corresponding weight matrix $\hat{W}_{u}^m$ of the same index. The size of each $\hat{W}_{u}^m$ is equal to the feature vector length $\times$ hidden layer width. The number of weight matrices is $2n+1$ (the $1$ is for the central frame), which matches the number of frames in the window. The output of a single CLNN step over a window of frames is a single representative vector.

Several CLNN layers can be stacked on top of each other to form a deep architecture as shown in Fig. \ref{fig:clnn}. The figure also depicts a number of $k$ extra frames remaining after the processing applied through the two CLNN layers. These $k$ extra frames allow incorporating an aggregation operation within the network by pooling the temporal dimension or they can be flattened to form a single vector before feeding them to a fully connected network. The CLNN is trained over segments following (\ref{eq:segments}) 
\vskip -0.25in
\begin{align}
\label{eq:segments}
q=(2n)m + k \qquad ,\,n,\,m\,and\,k \geq 1
\end{align}
\vskip -0.1in
where $q$ is the segment size, $n$ is the order, $m$ is the number of layers and $k$ is for the extra frames.   The input at each CLNN layer has $2n$ fewer frames than the previous layer. For example, for $n=4$, $m=3$ and $k=5$, the input is of size 29 frames. The output of the first layer is $29-2\times4=21$ frames. Similarly, the output of the second and third layers is $13$ and $5$ frames, respectively. The $5$ remaining frames of third layer are the extra frames to be pooled or flattened.
The activation at a hidden node of a CLNN can be formulated as in (\ref{eq:hidden_node_activation})
\vskip -0.25in
\begin{align}
\label{eq:hidden_node_activation}
y_{j,\; t} = f\left( b_{j} + \sum_{u=-n}^{n}\sum_{i=1}^{l} x_{i,\;u+t} \ W_{i,\;j,\;u}  \right)
\end{align}
\vskip -0.1in
where $y_{j,\;t}$ is the activation at node $j$ of a hidden layer for frame $t$ in a segment of size $q$. This frame is also the window's middle frame at $u=0$. The output $y$ is given by the value of the activation function $f$ when applied on the summation of the bias $b_j$ of node $j$ and the multiplication of $W_{i,\;j,\;u}$ and $x_{i,\;u+t}$. The input $x_{i,\;u+t}$ is the $i^{th}$ feature in a single feature vector of size $l$ at index $u+t$ within a window and $W_{i,\;j,\;u}$ is the weight between the $i^{th}$ input of a feature vector and the $j^{th}$ hidden node. The $u$ index (in $W_{i,\;j,\;u}$ and $x_{i,\;u+t}$) is for the temporal window of the interval of frames to be considered within $[-n+t,\;n+t]$. Reformulating (\ref{eq:hidden_node_activation}) in a vector form is given in (\ref{eq:hidden_vector}).
\vskip -0.25in
\begin{align}
\label{eq:hidden_vector}
\hat{y}_{t}=f\left(\hat{b} + \sum_{u=-n}^{n} \hat{x}_{u+t}  \cdot  \hat{W}_u  \right) 
\end{align}
\vskip -0.1in
where $\hat{y}_{t}$ is the activation vector observed at the hidden layer for the central frame conditioned on the input vectors in the interval $[\hat{x}_{-n+t}, \hat{x}_{n+t}]$ is given by the activation function $f$ applied on the summation of the bias vector $\hat{b}$ and the summation of the multiplication between the vector $\hat{x}_{u+t}$ at index $u+t$ ($t$ is for the window's middle frame at $u=0$ and the index of the frame in the segment) and the corresponding weight matrix $\hat{W}_u$ at the same index, where $u$ takes values in the range of the considered window from $-n$ up to $n$. The conditional distribution can be captured using a logistic function as in $p(\hat{y_{t}}|\hat{x}_{-n+t},...,\hat{x}_{-1+t},\hat{x}_{t},\hat{x}_{1+t},...,\hat{x}_{n+t}) = \sigma(...) $, where $\sigma$ is the hidden layer sigmoid function or the output layer softmax.   
\section{Masked Conditional Neural Networks}
\vskip -0.1in
\begin{figure}
\vskip -0.3in
\begin{center}
\centerline{\includegraphics[width=\textwidth]{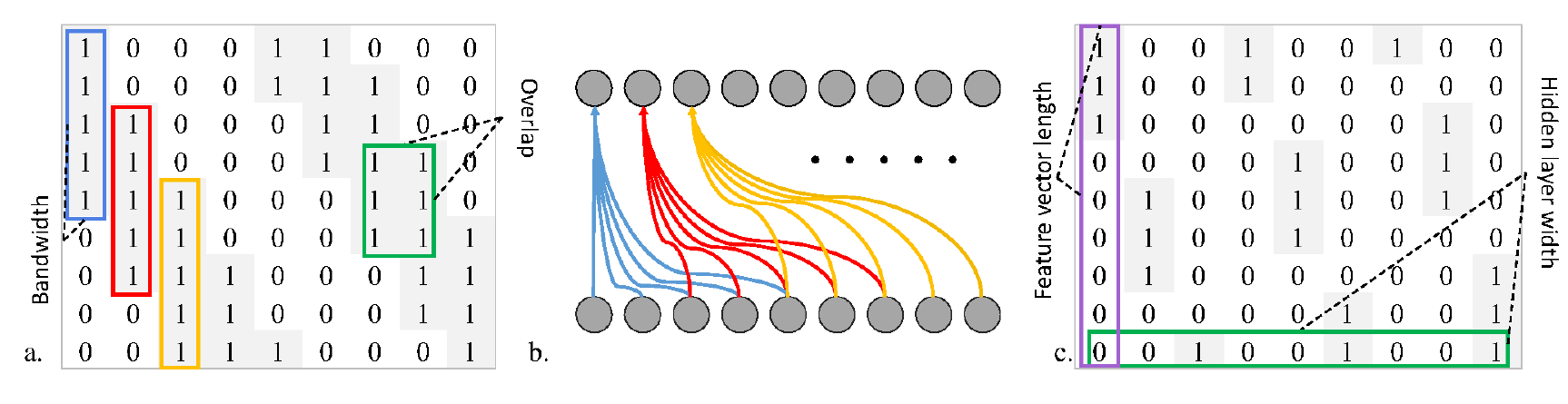}}
\vskip -0.1in
\caption[9pt]{Masking patterns. a) $Bandwidth = 5$ and $Overlap = 3$, b) the active links following the masking pattern in a. c) $Bandwidth = 3$ and $Overlap = -1$ }
\label{fig:mask}
\end{center}
\vskip -0.5in
\end{figure}
 
The Mel-Scaled analysis applied in MFCC and Mel-Scaled spectrograms, both used extensively as intermediate signal representation by sound recognition systems, exploit the use of a filterbank (a group of signal processing filters). Considering a sound signal represented in a spectrogram, the energy of a certain frequency bin may smear across nearby frequency bins. Aggregating the energy across neighbouring frequency bins is a possible representation to overcome the frequency shifts, which is tackled by filterbanks. More general mixtures across the bins could be hand-crafted to select the most prominent features for the signal under consideration. 

The Masked ConditionaL Neural Network (MCLNN), we introduce in this work embeds a filterbank-like behaviour and allows the exploration of a range of feature combinations concurrently instead of manually hand-crafting the optimum mixture of features. Fig. \ref{fig:mask} depicts the implementation of the filterbank-like behaviour through the binary mask enforced over the network's links that activate different regions of a feature vector while deactivating others following a band-like pattern. The mask is designed based on two tunable hyper-parameters: the bandwidth and the overlap. Fig. \ref{fig:mask}.a. shows a binary mask having a bandwidth of 5 (the five consecutive ones in a column) and an overlap of 3 (the overlapping ones between two successive columns). A hidden node will act as an expert in a localized region of the feature vector without considering the rest of it. This is depicted in Fig. \ref{fig:mask}.b. The figure shows the active connections for each hidden node over a local region of the input feature vector matching the mask pattern in Fig. \ref{fig:mask}.a. 
The overlap can be assigned negative values as shown in Fig. \ref{fig:mask}.c. The figure shows a mask with a bandwidth of 3 and overlap of $-1$, depicted by the non-overlapping distance between the 1's of two successive columns. Additionally, the figure shows an additional role introduced by the mask through the presence of shifted versions of the binary pattern across the first set of three columns compared to the second and third sets. The role involves the automatic exploration of a range of feature combinations concurrently. The columns in the figure map to hidden nodes. Therefore, for a single feature vector, the input at the $1^{st}$ node (corresponding to the $1^{st}$ column) will consider the first 3 features in the feature vector, the $4^{th}$  node will consider a different combination involving the first 2 features and the $7^{th}$ node will consider even a different combination using the first feature only. This behaviour embeds the mix-and-match operation within the network, allowing the hidden nodes to learn different properties through the different combinations of feature vectors meanwhile preserving the spatial locality. The position of the binary values is specified through a linear index $lx$ following  (\ref{eq:linearindex}) 

\vskip -0.25in
\begin{align}
\label{eq:linearindex}
lx=a + (g-1)(l+(bw-ov))
\end{align}
\vskip -0.1in

where $lx$ is given by bandwidth $bw$, the overlap $ov$ and the feature vector length $l$. The term $a$ takes the values in $[\;0, \; bw-1\;]$ and $g$ is in the interval $[  \;1, \lceil (l \times e)/(l+(bw-ov))\rceil\;  ]$.
The binary masking is enforced through an element-wise multiplication following (\ref{eq:mask}).
\vskip -0.25in
\begin{align}
\label{eq:mask}
\hat{Z}_u =\hat{W}_u \circ \hat{M}
\end{align}
\vskip -0.1in
where $\hat{W}_u$ is the original weight matrix at a certain index $u$ and $\hat{M}$ is the masking pattern applied. $\hat{Z}_u$ is the new masked weight matrix to replace the weight matrix in (\ref{eq:hidden_vector}).
\section{Experiments}
\vskip -0.1in
We performed the MCLNN evaluation using the GTZAN \cite{RN294} and the ISMIR2004 datasets widely used in the literature for benchmarking several MIR tasks including genre classification. 
The GTZAN consists of 1000 music files categorized across 10 music genres (blues, classical, country, disco, hip-hop, jazz, metal, pop, reggae and rock). The ISMIR2004 dataset comprise training and testing splits of 729 files each. The splits have 6 unbalanced categories of music genres (classical, electronic, jazz-blues, metal-punk, rock-pop and world) of full length recordings. All files were resampled at 22050 Hz and chunks of 30 seconds were extracted. Logarithmic Mel-Scaled 256 frequency bins spectrogram transformation was applied using an FFT window of 2048 ($\approx$ 100 msec) and an overlap of 50\%. The feature-wise z-score parameters of the training set was applied to both the validation and test sets. Segments of frames following (\ref{eq:segments}) were extracted. 
\begin{table}
    \begin{minipage}{.4\textwidth}
     \caption{Accuracies on the GTZAN}
  \label{table:gtzan}
      \centering
      \begin{tabular}{lc}
        \toprule \parbox[t]{4.7cm}{ Classifier and Features} & Acc.\%\\
        \midrule
CS + Multiple feat. sets\cite{RN317}$^2$&92.7\\
SRC + LPNTF + Cortical features\cite{RN318}$^2$&92.4\\
RBF-SVM + Scattering Trans.\cite{RN274}$^2$&91.4\\
\textbf{MCLNN + Mel$-$Spec.(this work)$^2$}&\textbf{85.1}\\
RBF-SVM + Spec.$-$DBN\cite{RN290}$^4$&84.3\\
\textbf{MCLNN + Mel$-$Spec.(this work)}$^3$&\textbf{84.1}\\
Linear SVM + PSD on Octaves\cite{RN291}$^3$&83.4\\
Random Forest + Spec.$-$DBN\cite{RN284}$^5$& 83.0\\
AdaBoost + Several features\cite{RN295}$^1$&83.0\\
RBF SVM + Spectral Covar.\cite{RN321}$^2$&81.0\\
Linear SVM + PSD on frames\cite{RN291}$^3$&79.4\\
SVM + DWCH\cite{RN319}$^2$&78.5\\
        \bottomrule
      \end{tabular}
    \end{minipage}
    \begin{minipage}{0.7\textwidth}
     \caption{Accuracies on the ISMIR2004}
  \label{table:ismir}
      \centering
      \begin{tabular}{lc}
        \toprule \parbox[t]{4.7cm}{Classifier and Features} & Acc.\%\\
        \midrule 
SRC + NTF + Cortical features\cite{RN318}$^9$&94.4\\
KNN + Rhythm\&timbre\cite{RN343}$^2$&90.0\\
SVM + Block-Level features \cite{RN333}$^8$&88.3\\
\textbf{MCLNN + Mel$-$Spec.(this work)}$^2$&\textbf{86.0}\\
\textbf{MCLNN + Mel$-$Spec.(this work)}$^3$&\textbf{84.8}\\
\textbf{MCLNN + Mel$-$Spec.(this work)}$^9$&\textbf{84.8}\\
GMM + NMF\cite{RN324}$^1$&83.5\\
\textbf{MCLNN + Mel$-$Spec.(this work)}$^6$&\textbf{83.1}\\
SVM + Symbolic features \cite{RN341}]$^2$&81.4\\
NN + Spectral Similarity FP \cite{RN338}$^7$&81.0\\
SVM + High-Order SVD \cite{RN337}$^2$&81.0\\
SVM + Rhythm and SSD \cite{RN339}$^6$&79.7\\
        \bottomrule
      \end{tabular}
    \end{minipage}
 
 \scalebox{0.795}{
\begin{tabular}{lll}
\addlinespace[0.1cm]
$^{1}$5-fold cross-validation&$^{4}$50\% training, 20\% validation and 30\% testing&$^{7}$leave-one-out cross-validation\\
$^{2}$10-fold cross-validation&    $^{5}$4$\times$50\% training, 25\% validation and 25\% testing&$^{8}$Not referenced\\
$^{3}$10$\times$ 10-fold cross-validation&$^{6}$10$\times$(Train 729 file , test 729 file)&$^{9}$Train 729 files,test 729 files \\
\end{tabular}}
\vskip -0.1in
\end{table}

\begin{table}

    \begin{minipage}{.4\textwidth}
       \centering
       
       \caption{MCLNN parameters}
  	   \label{table:modelparam}
  	   \vskip -0.2cm
       \begin{tabular}{lcccc}
\hline
\addlinespace[0.1cm]
\parbox[t]{1cm}{\multirow{2}{*}{\centering Layer}}  & \parbox[t]{1cm}{\centering Hidden \\ Nodes} & \parbox[t]{1cm}{\centering MCLNN \\ Order} &  \parbox[t]{1.5cm}{\centering Mask \\ Bandwidth}   & \parbox[t]{1.2cm}{\centering Mask \\ Overlap}   \\[4ex]
\addlinespace[-0.1cm]
\hline
\addlinespace[0.1cm]
1    & 220 & 4 & 40 & -10 \\
2    & 200 & 4 & 10 & 3 \\
\hline
		\end{tabular}
	\end{minipage}
    \begin{minipage}{0.72\textwidth}
      \centering
      \caption{GTZAN random and filtered}
  	  \label{table:gtzanbob}
  	  \vskip -0.2cm
      \begin{tabular}{lcc}
\hline
\addlinespace[0.1cm]
\parbox[t]{1cm}{\multirow{2}{*}{\centering Model}} & \parbox[t]{1.2cm}{\centering Random Acc. \%} & \parbox[t]{1.2cm}{\centering Filtered Acc. \%}    \\[4ex]
\addlinespace[-0.1cm]
\hline
\addlinespace[0.1cm]
\textbf{MCLNN(this work)} &\textbf{84.4}&\textbf{65.8}  \\
DNN [25]&81.2&42.0  \\
\hline
		\end{tabular}
	\end{minipage}
\vskip -0.27in
\end{table}

The network was trained to minimize the categorical cross entropy between the segment's predicted label and the target one. The final decision of the clip's genre is decided based on a majority voting across the frames of the clip. The experiments for both datasets were carried out using a 10-fold cross-validation that is repeated for 10 times. An additional experiment was applied using the ISMIR2004 dataset original split (729 training, 729 testing) that was also repeated for 10 times. 
We adapted a two-layered MCLNN, as listed in Table \ref{table:modelparam}, followed by a single dimensional global mean pooling \cite{RN332} layer to pool across $k=10$ extra frames and finally a 50 node fully connected layer before the softmax output layer. Parametric Rectified Linear Units (PReLU) \cite{RN283} were used for all the model's neurons. We applied the same model to both datasets to gauge the generalization of the MCLNN to datasets of different distributions.
Tables \ref{table:gtzan} and Table \ref{table:ismir} list the accuracy achieved by the MCLNN along with other methods widely cited in the literature for the genre classification task on the GTZAN and the ISMIR2004 datasets. MCLNN surpasses several state-of-the-art methods that are dependent on hand-crafted features or neural networks, achieving an accuracy of 85.1\% and 86\% over a 10-fold cross-validation for the GTZAN and ISMIR2004, respectively. We repeated the 10-fold cross-validation 10 times to validate the accuracy stability of the MCLNN, where the MCLNN achieved 84.1\% and 84.83\% over the 100 training runs for each of the GTZAN and the ISMIR2004, respectively. 

To further evaluate the MCLNN performance, we adapted the publicly available splits released by Kereliuk et al.\cite{RN316}. In their work, they released two versions of splits for the GTZAN files: a randomly stratified split (50\% training, 25\% validation and 25\% testing) and a fault filtered version, where they cleared out all the mistakes in the GTZAN as reported by Sturm \cite{RN313}, e.g. repetitions, distortion, etc. As listed in Table \ref{table:gtzanbob}, MCLNN achieved 84.4\% and 65.8\% compared to Kereliuk's attempt that achieved 81.2\% and 42\% for the random and fault-filtered, respectively, in their attempt to reproduce the work by Hamel et al. \cite{RN290}.
The experiments show that MCLNN performs better than several neural networks based architectures and comparable to some other works dependent on hand-crafted features. MCLNN achieved these accuracies irrespective of the rhythmic and perceptual properties \cite{RN428} that were used by methods that reported higher accuracies than the MCLNN. 
Finally, we wanted to tackle the problem of the data size used in training, referring to the works in \cite{RN290,RN295,RN291,RN321,RN316}, an FFT window of 50 msec was used. On the other hand, the MCLNN achieved the mentioned accuracies using a 100 msec window, which decreases the number of feature vectors to be used in classification by 50\% and consequently the training complexity, which allows the MCLNN to scale for larger datasets.
\section{Conclusions and Future work}
\vskip -0.1in
We have introduced the ConditionaL Neural Network (CLNN) and its extension the Masked ConditionaL Neural Network (MCLNN). The CLNN is designed to exploit the properties of the multi-dimensional temporal signals by considering the sequential relationship across the temporal frames. The mask in the MCLNN enforces a systematic sparseness that follows a frequency band-like pattern. Additionally, it plays the role of automating the the exploration of a range of feature combinations concurrently analogous to the exhaustive manual search for the hand-crafted feature combinations. We have applied the MCLNN to the problem of genre classification. Through an extensive set of experiments without any especial rhythmic or timbral analysis, the MCLNN have sustained accuracies that surpass neural based and several hand-crafted feature extraction methods referenced previously on both the GTZAN and the ISMIR2004 datasets, achieving 85.1\% and 86\%, respectively. Meanwhile, the MCLNN still preserves the generalization that allows it to be adapted for any temporal signal. Future work will involve optimizing the mask patterns, considering different combinations of the order across the layers. We will also consider applying the MCLNN to other multi-channel temporal signals.

\subsubsection*{Acknowledgments.} This work is funded by the European Union's Seventh Framework Programme for research, technological development and demonstration under grant agreement no. 608014 (CAPACITIE).
\begin{ssmall}
\bibliography{referencefile}
\end{ssmall}

\end{document}